\documentclass[10pt]{article}
\usepackage[utf8]{inputenc} % allow utf-8 input
\usepackage[T1]{fontenc}    % use 8-bit T1 fonts
\usepackage{hyperref}       % hyperlinks
\usepackage{url}            % simple URL typesetting

\usepackage{graphicx}

\usepackage{comment}
\usepackage{color}
\usepackage{dsfont}

\usepackage{wrapfig}

\usepackage[body={6.5in,8in}]{geometry}

\usepackage{multirow}
\usepackage{multicol}
\usepackage{times}
\usepackage{epsfig}
\usepackage{graphicx}
\usepackage{amsmath}
\usepackage{amssymb}
\usepackage[ruled,vlined,linesnumbered]{algorithm2e}
\usepackage{booktabs}
\usepackage{array}
\usepackage{setspace}
\usepackage{caption}

\newtheorem{theo}{Theorem}

\date{}

\author{
Hongxin Wei, Lei Feng, Rundong Wang, Bo An\\
  Nanyang Technological University, Singapore\\

  \texttt{\{hongxin001,feng0093,rundong001\}@e.ntu.edu.sg}, \texttt{\{boan\}@ntu.edu.sg}\\
}

\begin{document}

%%%%%%%%% TITLE
\title{MetaInfoNet: Learning Task-Guided Information for Sample Reweighting}

\maketitle

%%%%%%%%% ABSTRACT
\begin{abstract}
Deep neural networks have been shown to easily overfit to biased training data with label noise or class imbalance. 
%It was shown that deep neural networks could easily overfit to biased training data with label noise or class imbalance.
Meta-learning algorithms are commonly designed to alleviate this issue in the form of sample reweighting, by learning a meta weighting network that takes training losses as inputs to generate sample weights. In this paper, we advocate that choosing proper inputs for the meta weighting network is crucial for desired sample weights in a specific task, while training loss is not always the correct answer. %For instance, meta network trained with training loss works well under uniform noise but performs bad under flipping noise. %the situation is quite the opposite when trained with logits as input. 
In view of this, we propose a novel meta-learning algorithm, MetaInfoNet, which automatically learns effective representations as inputs for the meta weighting network by emphasizing task-related information with an information bottleneck strategy.
%MetaInfoNet does not require hand-crafted inputs or a specified loss function, thereby 
%As a result, MetaInfoNet allows to apply meta sample re-weighting methods to achieves promising results in various tasks, without carefully designed inputs.
Extensive experimental results on benchmark datasets with label noise or class imbalance validate that MetaInfoNet is superior to many state-of-the-art methods.

%limited information can be included in loss value as a simple scalar, making it hard to learn a effective weight. 
\end{abstract}

%%%%%%%%% BODY TEXT
\section{Introduction}
Deep Neural Networks (DNNs) have achieved remarkable success on various computer vision tasks due to their powerful capacity for modeling complex input patterns. Despite their success, the vulnerability of DNNs has been extensively illustrated in many previous studies \cite{arpit2017closer,buda2018systematic, galar2011review, kawaguchi2017generalization, neyshabur2017exploring, novak2018sensitivity}.
One important drawback of DNNs is that DNNs could easily overfit to biased training data, where the distribution of training data is inconsistent with that of the evaluation data.

There are many different forms of distribution mismatch, leading to poor performance in generalization. A typical example is class imbalance \cite{cui2018large, huang2016learning}, where the distribution of data across the classes is not equal in the training set. This issue will sometimes lead to biased training models that does not perform well in practice \cite{bengio2015sharing, ouyang2016factors,van2017devil}. Another popular type of distribution mismatch is label noise, which usually happens when the training set is collected from a crowdsourcing system \cite{yan2014learning} or search engines \cite{blum2003noise}. It has been shown that a standard CNN can fit any ratio of label noise in the training set and eventually leads to poor generalization performance \cite{arpit2017closer, zhang2018generalized}. Therefore, robust learning from these biased data has become an important and challenging problem in machine learning and computer vision.

\begin{figure}[!t]
\centering
\includegraphics[scale=0.5]{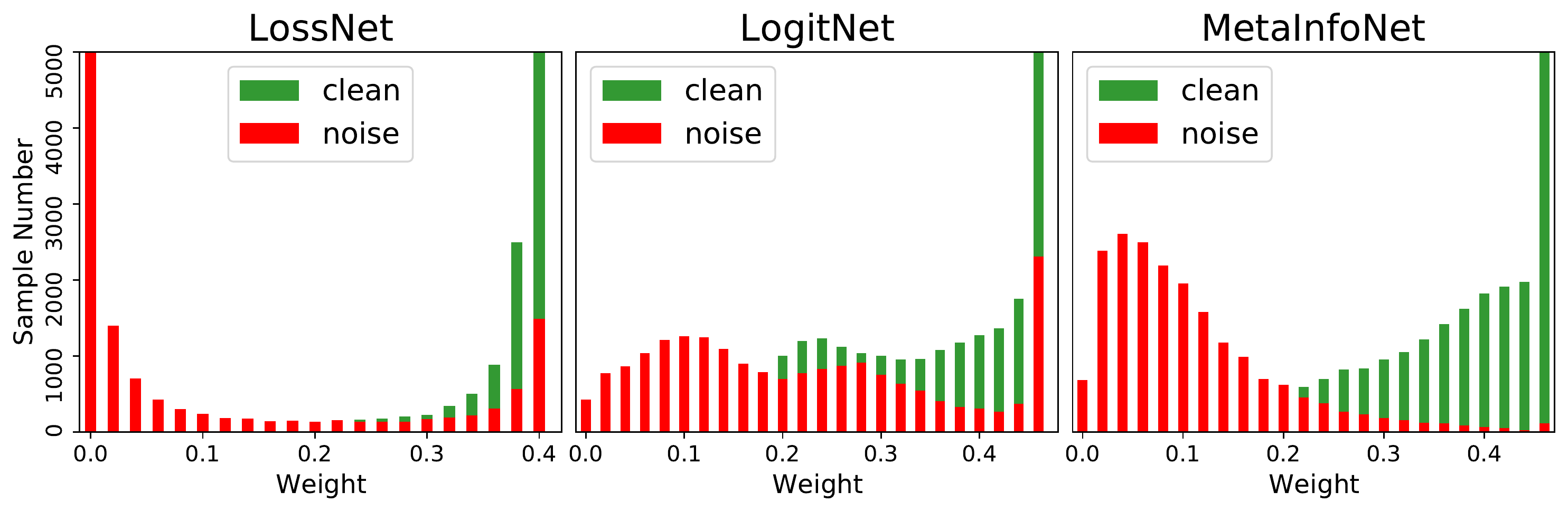}
\caption{Sample weight distributions on training
data of different instatiations (different inputs) of meta weighting network under the 40\% Flip-1 label noise case. It is ideally expected that higher weights are assigned to clean data (green) while lower weights are assigned to noisy data (red).
%Generally, we expect to allocate low weight for the samples with noisy labels, and high weight for the samples with clean labels.
}
\label{fig:sample_weight}
\end{figure}

Sample reweighting algorithms \cite{kumar2010self,lin2017focal,malisiewicz2011ensemble} are widely studied for robust learning from biased data. The main idea is to generate different weights for training losses to different samples. Some existing algorithms design specific weighting function with training loss by simple rules, such as monotonically increasing \cite{lin2017focal,malisiewicz2011ensemble} or monotonically decreasing \cite{de2003framework, kumar2010self, zhang2018generalized}, which means taking samples with larger or smaller loss values as more important ones. However, these methods need to manually design a specific form of weighting function based on certain assumptions on training data. To make the learning more automatic and reliable across various biased settings, a popular research line is meta sample weighting, which learns weights for each sample implicitly \cite{ren18l2rw} or automatically learns an explicit weighting function \cite{shu2019meta}. Specifically, the weighting function could be parameterized as a multilayer perceptron (MLP) network that maps the training loss to sample weight \cite{shu2019meta}, and the training of the parameters could by guided by a small unbiased validation set. For easy reference, we call this instantiation of meta weighting network (MW-Net) as \textbf{LossNet} \cite{shu2019meta}. 

In this paper, we advocate that choosing proper inputs for MW-Net is crucial for desired sample weights in a specific task, while training loss is not always the correct answer. As shown in Figure \ref{fig:sample_weight}, we illustrate the sample weight distributions on training data of different instantiations (different inputs) of MW-Net. We can observe that LossNet tends to produce polarized results, and high weights could be assigned to some noisy data. This is because the commonly used cross-entropy loss is known to be highly overconfident \cite{menon2019can,thulasidasan2019mixup}. Besides, the training loss may not contain rich and flexible information of the original sample for capturing meaningful sample weights, due to the restriction on the fixed loss function.

%Despite the effectiveness of LossNet, the learned loss-weight mapping is too simple and little information is included in the training loss as a single scalar. On the other hand, the commonly used cross-entropy loss is known to be highly overconfident \cite{thulasidasan2019mixup,menon2019can}, thereby makes LossNet tend to output extreme values to weight samples. The phenomenon is explicitly shown in the left of Figure \ref{fig:sample_weight}.   %Although LossNet performs well on some of bias settings like uniform noise, it is not efficient in other setting such as flip-1 noise, as shown in Figure \ref{fig:compare_cifar10}. %Although the MLP network is theoretically a universal approximator for almost any continuous function, %We show that choosing a proper input is the key to train an efficient meta network for a specific task. 
To alleviate this issue, an intuitive method is to use the logits and labels as inputs (instead of training loss with a specified loss function) to build MW-Net (we call such a instantiation of MW-Net as \textbf{LogitNet}). As shown in Figure \ref{fig:sample_weight}, LogitNet would not produce polarized results, while it could still assign high sample weights to some noisy data. This is because LogitNet may contain too much redundant information with the unprocessed logits.
%However, although richer information is embodied in such inputs, LogitNet may overfit to the task-independent information due to the small size of the meta dataset, as shown in the center of Figure \ref{fig:sample_weight}. 
To address this problem, we propose a novel meta-learning algorithm, \textbf{MetaInfoNet}, which automatically learns effective representations as inputs for MW-Net by emphasizing task-related information with an information bottleneck strategy. As shown in Figure \ref{fig:sample_weight}, our proposed MetaInfoNet can learn smooth and meaningful sample weights, and the trend that higher weights are assigned to clean data while lower weights are assigned to noisy data is clearly demonstrated.

Finally, we conduct extensive experiments on simulated and real-world datasets with class imbalance or label noise, to show that MetaInfoNet significantly improves the robustness of deep learning on training data under various biased settings. Empirical results demonstrate that the robustness of deep models trained by our proposed MetaInfoNet is superior to many state-of-the-art methods.

% In summary, the key contributions of this paper are as follows:
% \begin{itemize}
%     \item We propose MetaInfoNet, a novel algorithm that learning a inputs which is sufficient for from logits, the given labels
    
%     \item 
    
%     \item For practical performance, we conduct extensive experiments on synthesized 

% \end{itemize}

\section{Related Work}

\noindent\textbf{Sample Reweighting Methods.} 
The idea of sample reweighting has been commonly used in the machine learning literature. %The main idea is to generate weights for losses of samples in the training process. Some existing algorithms design specific weighting function with training loss by simple rules, such as monotonically increasing \cite{malisiewicz2011ensemble, lin2017focal} or monotonically decreasing \cite{de2003framework, kumar2010self, zhang2018generalized}, which means taking samples with larger or smaller loss values as more important ones. 
For example, hard example mining downsamples the majority class and exploits the most challenging examples \cite{malisiewicz2011ensemble}. Similarly, Focal loss emphasizes harder examples by soft weighting \cite{lin2017focal}. On the contrary, self-paced learning (SPL) takes samples with smaller loss values as more important ones firstly \cite{kumar2010self}. Despite their success under some specific settings, these methods need to manually design a particular form of weighting function based on certain assumptions on training data, which might be impractical. Rather than predefined by human experts, MentorNet uses a bidirectional LSTM network to learn a curriculum from data with label noises \cite{jiang2018mentornet}. However, the weighting function of MentorNet is too complicated and would overfit to the biased data.

\noindent\textbf{Meta Learning Methods for Robustness.}
 %To make the learning more automatic and reliable across various biased settings, some 
Meta-learning algorithms are introduced to improve robustness of deep learning in the form of sample reweighting. The first work is L2RW \cite{ren18l2rw}, which implicitly learns the weights without a pre-defined weighting function, then uses a small unbiased validation set to guide the training of its parameters. %This, however, might lead to unstable weighting behavior during training and unavailability for generalization. 
 In contrast, LossNet \cite{shu2019meta} parameterizes the weighting function as an MLP network explicitly, mapping from training loss to sample weight. %Although the MLP network is theoretically a universal approximator for almost any continuous function, 
 However, the capacity of LossNet is limited by its input with the fixed loss function.

\noindent\textbf{Learning with Class Imbalance.}
In addition to sample reweighting, there are other methods to handle the class imbalance issue in deep learning. For example, some methods try to transfer the knowledge learned from major classes to minor classes \cite{cui2018large, wang2017learning}. The metric learning based methods have also been developed to effectively exploit the tailed data to improve the generalization ability, e.g., triple-header loss \cite{huang2016learning} and range loss \cite{zhang2017range}.

\noindent\textbf{Learning with Label Noise.}
For handling noisy label issues \cite{feng2020can,han2019deep,li2020dividemix,ma2020normalized,nguyen2019self,tanaka2018joint,wang2019imae, wang2019symmetric}, some other algorithms focus on estimating the label transition matrix. For example, F-correction \cite{patrini2017making} proposed a loss correction approach by heuristically estimating the noise transition matrix. In these approaches, the quality of noise rate estimation is a critical factor for improving robustness. However, noise rate estimation is challenging, especially on datasets with a large number of classes. Another popular research line of handling noisy labels is to train models on small-loss instances, which can be viewed as a hard version of sample weighting \cite{han2018co,jiang2018mentornet,wei2020combating}.

\noindent\textbf{Information Bottleneck.} Information Bottleneck (IB) method was initially proposed in \cite{tishby2000information}, where the idea can be formulated as a variational principle of minimizing the mutual information between the input and the learned representation, while preserving the information about the learning task. The IB method has been successfully applied to supervised learning \cite{alemi2016deep}, generative modeling \cite{peng2018variational} and reinforcement learning \cite{peng2018variational, wang2019learning}. In this work, we apply the IB algorithm to enforce the input of MW-Net to focus on the relevant information implicitly defined by the learning task.

\section{Preliminaries}
In this section, we introduce the problem setting of learning from biased training data with sample reweighting and the formulation of LossNet \cite{shu2019meta}, which is a representative related work.

\subsection{Problem Setting}

For multi-class classification with $K$ classes, we suppose the biased training dataset with $N$ samples is given as $\mathcal{D}^{\mathrm{train}} = \{\boldsymbol{x}_i, y_i\}^N_{i=1}$, where $\boldsymbol{x}_i$ is the $i$-th instance with its observed label as $y_i \in \{1, \dots, K\}$. Similar to L2RW and LossNet, we assume that there is a small unbiased and clean meta dataset with $M$ samples $\mathcal{D}^{\mathrm{val}} = \{x^{\mathrm{val}}_i, y^{\mathrm{val}}_i\}^M_{i=1}$ and $M \ll N$, representing the meta-knowledge of ground-truth sample-label association. For the classifier to be trained, we denote it as $f_{\Theta}(\boldsymbol{x})$, where $\Theta$ is the parameters of the classifier. Generally, we optimize the classifier network by minimizing the training loss: $\frac{1}{N} \sum^N_{i=1} L^{\mathrm{train}}_i(\Theta) = \frac{1}{N} \sum^N_{i=1} \ell(f_{\Theta}(\boldsymbol{x}_i), y_i)$, where $\ell$ denotes the employed loss function (e.g., cross entropy loss), and each input example is weighted equally. For enhancing the robustness of training on the biased training data, we aim to assign weight $w_i$ on the loss of the $i$-th sample. Without loss of generality, the weighting function can be formulated as $\mathcal{V}_{\Phi}((\boldsymbol{x}_i, y_i), f_{\Theta})$ where $\Phi$ denotes the parameters of the weighting function. 
%Here, we minimize a weighted loss: $L^{train}(w) = \sum^N_{i=1} w_i L^{train}_i $. 
Then the optimal parameter $\Theta$ of classifier is calculated by minimizing the following weighted loss:
\begin{equation}
\label{eq:training_loss}
\begin{split}
\Theta^{*}(\Phi) &= \underset{\Theta}{\arg\min} \  \mathcal{L}^{\mathrm{train}}(\Theta;\Phi)\\
&\triangleq\underset{\Theta}{\arg\min} \frac{1}{N} \sum_{i=1}^{N} w_i L^{\mathrm{train}}_i(\Theta),
\end{split}
\end{equation}
where $w_i = \mathcal{V}((\boldsymbol{x}_i, y_i), f_{\Theta}; \Phi)$ denotes the example weight and $L^{\mathrm{train}}_i(\Theta) = \ell(f_{\Theta}(\boldsymbol{x}_i), y_i)$ denotes the loss on the training example $(\boldsymbol{x}_i,y_i)$.

\subsection{Formulation of LossNet}

\begin{figure}[!t]
\centering
\includegraphics[scale=0.4]{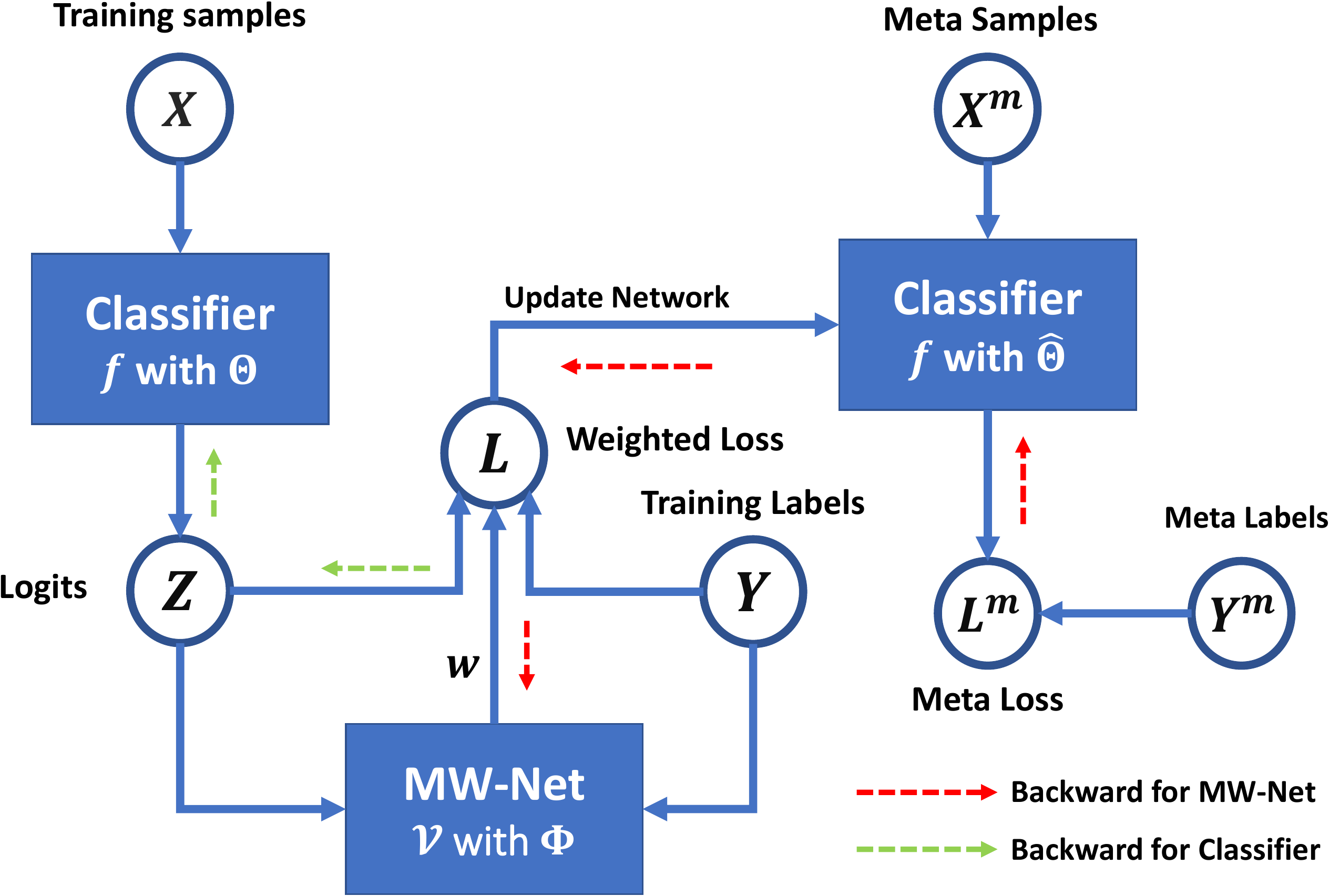}
\caption{Framework for Meta Weighting.}
\label{fig:network}
\end{figure}

LossNet \cite{shu2019meta} formulates the weighting function $\mathcal{V}$ as an MLP network, and automatically trains it in a meta-learning manner. Specifically, the MLP network learns a mapping from training loss to sample weight, $w_i = \mathcal{V}_{\Phi}(L_i^{\mathrm{train}}(f_{\Theta}(\boldsymbol{x}_i), y_i))$, where $\Phi$ denotes the parameters of the MLP net. In the training process, the optimal parameter $\Phi^{*}$ of LossNet is calculated by minimizing the following meta loss:
\begin{equation}
\label{eq:meta_loss}
\begin{split}
\Phi^{*} &= \underset{\Phi}{\arg\min} \  \mathcal{L}^{\mathrm{meta}}(\Theta^*(\Phi))\\
&\triangleq\underset{\Phi}{\arg\min}  \frac{1}{M} \sum_{i=1}^{M} L_{i}^{\mathrm{meta}}(\Theta^*(\Phi)),
\end{split}
\end{equation}
where $L_{i}^{\mathrm{meta}}(\Theta^*) = \ell(f_{\Theta^*}(x^{\mathrm{val}}_i), y_i^{\mathrm{val}}))$ denotes the meta loss on the validation example $(\boldsymbol{x}^{\mathrm{val}}_i,y^{\mathrm{val}}_i)$.
%\noindent where $L_{i}^{\mathrm{meta}}(\Theta) = L_{ce}(f_{\Theta}(x^{m}_i), y_i^{m})$ is calculated on the meta dataset.

As we analyzed before, the training loss may not  contain rich and flexible information of the original sample for capturing meaningful sample weights, due to the restriction on the fixed loss function. As the commonly used cross-entropy loss is known to be highly overconfident \cite{menon2019can,thulasidasan2019mixup}, LossNet tends to produce polarized results, and high weights could be assigned to some noisy data.

%LossNet relies on a key assumption: there is sufficient information in the training losses, for proper weighting the samples under various biased settings. %This assumption might be impractical when the loss is not calibrated. On one hand, the information in training loss, a single scalar, is limited and cannot be extended, thereby narrowing the range of applicable biased settings.
%For instance, LossNet performs badly under the flip-40\% case.
%todo: describe why not be met? On the other hand, we need to ask whether the losses maintain \textit{classification calibration} \cite{menon2019can}. This is a minimum requirement on a loss to be useful for classification \cite{bartlett2006convexity,zhang2004statistical}. However, the commonly used cross-entropy loss is known to be overconfidence in the training of deep neural networks \cite{thulasidasan2019mixup,menon2019can}.   

%Specifically, in our experiments, we show that choosing a proper input is the key to train an efficient meta network for a specific task and training loss is not always the answer. % the situation is quite the opposite when trained with logits as input.

\section{The Proposed Approach}
In this section, we propose to automatically learn proper inputs for MW-Net, thereby adapting to various biased settings. %todo: add some content about advantage.

\begin{algorithm}[t]
\caption{The Meta Weighting Algorithm}
\label{alg:meta}
\SetAlgoLined
\setstretch{1.1}
\KwData{Training dataset $\mathcal{D}$, meta dataset $\mathcal{D}_m$, batch size $n,m$, max iterations $T$.}
\KwResult{Parameter $\Theta$ of classifier $f$.}
 Initialize parameters of Classifier and MW-Net\;

 \For{$t=0$ $\mathrm{to}$ $T$}{
    $\{\boldsymbol{x}_i,y_i\}_{i=1}^n \leftarrow \text{SampleMiniBatch}(\mathcal{D},n)$\;
    $\{x^{\mathrm{val}}_i,y^{\mathrm{val}}_i\}_{i=1}^m \leftarrow \text{SampleMiniBatch}(\mathcal{D}^{\mathrm{val}},m)$\;
    $ \boldsymbol{z}_i = f_{\Theta}(\boldsymbol{x}_i), \tilde{\boldsymbol{w}} = \mathcal{V}_{\Phi^{(t)}}(\boldsymbol{z},y)$\;
    $w_i \leftarrow \frac{\tilde{w}_i}{\sum_{j} \tilde{w}_j +\delta\left(\sum_{j} \tilde{w}_j\right)}$ \;
    $\mathcal{L}^{\mathrm{train}} = \frac{1}{n} \sum_{i=1}^{n} w_i \cdot\ell(\boldsymbol{z}_i, y_i)$\;
    $\nabla \Theta \leftarrow \text { BackwardAD }\left(\mathcal{L}^{\mathrm{train}}, \Theta\right)$\;
    $\widehat{\Theta} \leftarrow \Theta-\alpha \nabla \Theta$\;
    $\mathcal{L}^{\mathrm{meta}} = \frac{1}{m} \sum_{i=1}^{m} \ell(f_{\widehat{\Theta}}(x^{\mathrm{val}}_i), y^{\mathrm{val}}_i)$ \;
    Update MW-Net $\mathcal{V}$ to $\Phi^{(t+1)}$ by $\mathcal{L}^{\mathrm{meta}}$ \;
    $\tilde{\boldsymbol{w}} = \mathcal{V}_{\Phi^{(t+1)}}(\boldsymbol{z},y); w_i \leftarrow \frac{\tilde{w}_i}{\sum_{j} \tilde{w}_j+\delta\left(\sum_{j} \tilde{w}_j\right)}$ \;
    $\mathcal{L}^{\mathrm{train}} = \frac{1}{n} \sum_{i=1}^{n} w_i L^{\mathrm{train}}_i(\boldsymbol{z}_i, y_i)$ \;
    % $L = \frac{1}{N} \sum_{i=1}^{N} w L_{ce}(\widehat{y}, y)$\;
    Update Classifier $f$ to $\Theta^{(t+1)}$ by $\mathcal{L}^{\mathrm{train}}$ \;
 }
\end{algorithm}

\subsection{A General Framework for Meta Weighting}
\begin{figure}[!t]
\centering
\includegraphics[scale=0.35]{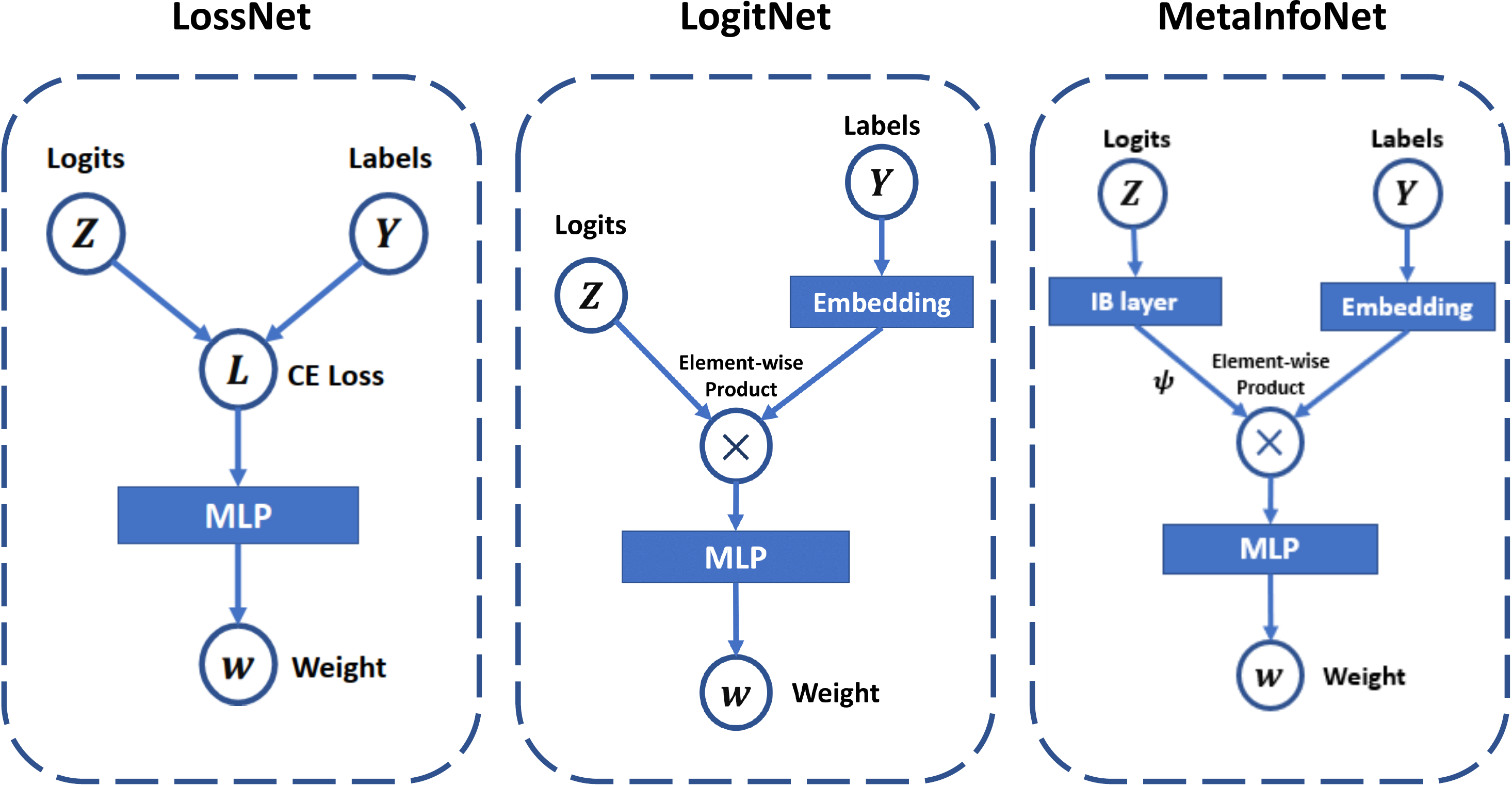}
\caption{Three different instantiations of MW-Net, including LossNet, LogitNet, and MetaInfoNet.}
\label{fig:compare}
\end{figure}

As shown in Figure \ref{fig:network}, we define a general framework for meta weighting, which contains a MW-Net that takes information from the samples $\boldsymbol{x}_i$ and its given labels $y_i$ (instead of training loss) as inputs. To abstract information from the sample $\boldsymbol{x}_i$, we adopt the outputs of the classifier $f$, (i.e., $f_{\Theta}(\boldsymbol{x}_i)$) for MW-Net. In this way, LossNet can be interpreted as an instantiation of MW-Net. Intuitively, if we calculate the cross-entropy loss with the logit and its corresponding label, we can exactly recover the LossNet model. The training loss can be seen as a representation of information abstracted from samples $\boldsymbol{x}_i$ and its given labels $y_i$. The left of Figure \ref{fig:compare} shows that LossNet is an instantiation of MW-Net. Under the general framework in Figure \ref{fig:network}, LossNet could be replaced by a more powerful instantiation of MW-Net, thereby achieving better performance.
%easily generalized and extended by enriching the representation of the inputs. 

Similar to L2RW \cite{ren18l2rw} and LossNet \cite{shu2019meta}, calculating the optimal parameters $\Theta^*$ and $\Phi^*$ requires two nested loops of optimization. To improve the optimization efficiency, we adopt an online strategy to alternatively update the classifier $f$ with $\Theta$ and the MW-Net $\mathcal{V}$ with $\Phi$ through a single optimization loop. We describe the details of the training process in Algorithm \ref{alg:meta}. In general, the training process can be separated into three parts:

\noindent\textbf{Virtually updating parameters of classifier.} In the first part, we randomly sample a mini-batch of training samples ${(\boldsymbol{x}_i, y_i), 1 \leq i \leq n}$ from the training dataset, where $n$ is the mini-batch size. Then the virtual updating of the classifier network parameter can be formulated by moving the current $\Theta^{(t)}$ along the descent direction of the objective loss in Eq. (\ref{eq:training_loss}) on a mini-batch training data (lines 5-9) :
\begin{equation}
\nonumber
\widehat{\Theta}^{(t)}(\Phi) = \Theta^{(t)} - \alpha \frac{1}{n} \left. \sum_{i=1}^{n} \mathcal{V}_{\Phi}((f_{\Theta}(\boldsymbol{x}_i), y_i)) \nabla_{\Theta} L^{\mathrm{train}}_{i}\right|_{\Theta^{(t)}},
\end{equation}
\noindent where $\alpha$ is the step size.

\noindent\textbf{Updating parameters of MW-Net.} After receiving the feedback of the virtually updated classifier, i.e., $\widehat{\Theta}^{(t)}(\Phi)$, the parameter $\Phi$ of the MW-Net can then be readily updated with Eq. (\ref{eq:meta_loss}) calculated on the meta dataset (lines 10-11): %i.e., moving the current parameter $\Phi^{t}$ along the objective gradient of Eq. \ref{eq:meta_loss} calculated on the meta-data 

\begin{equation}
\label{eq:update_mw}
\Phi^{(t+1)}=\Phi^{(t)}-\beta \frac{1}{m} \left.\sum_{i=1}^{m} \nabla_{\Phi} L_{i}^{\mathrm{meta}}(\widehat{\Theta}^{(t)}(\Phi))\right|_{\Phi^{(t)}},
\end{equation}
\noindent where $\beta$ is the step size.

\noindent\textbf{Updating parameters of classifier.} Then, the updated $\Phi^{(t+1)}$ is employed to ameliorate the parameter $\Theta$ of the classifier (lines 12-14), i.e.,
\begin{align}
\nonumber
&\Theta^{(t+1)}=\Theta^{(t)}\\
\nonumber
&\quad\quad\quad\quad-\alpha \frac{1}{n}\left.\sum_{i=1}^{n} \mathcal{V}_{\Phi^{(t+1)}}\left((f_{\Theta}(\boldsymbol{x}_i), y_i)\right) \nabla_{\Theta} L_{i}^{\mathrm{train}}\right|_{\Theta^{(t)}},
\end{align}

\subsection{LogitNet: An Instatiation of MW-Net}
Now the question becomes how to design MW-Net, for mapping information from the sample and its given label to weight. An intuitive idea is to simply take the model output of $\boldsymbol{x}_i$, i.e., $f_{\Theta}(\boldsymbol{x})$. For the given label $y_i$, we employ a label embedding layer to produce an embedding vector $\boldsymbol{e}(y_i)\in\mathbb{R}^K$, which has the same dimension as $f_{\Theta}(\boldsymbol{x})$. It is a fully connected layer that projects the given label to a dense vector. The obtained embedding vector can be seen as a latent vector for describing the properties of a specific class, in the context of MW-Net.

Since MW-Net adopts two pathways to extract information from $\boldsymbol{x}_i$ and $y_i$, it is intuitive to combine the vectors from the two pathways (i.e., $f_{\Theta}(\boldsymbol{x})$ and $\boldsymbol{e}(y_i)$) by concatenating them. However, a vector concatenation cannot account for any interactions between samples and its given labels. Therefore, we apply an element-wise product to combine them as the input for an MLP in MW-Net, that is, $w_i=\text{MLP}(f_{\Theta}(\boldsymbol{x}_i) \odot \boldsymbol{e}(y_i))$,
where $\odot$ denotes the element-wise product. In this way, the combined vector can capture the interaction information between the sample and its given label before we feed it into an MLP. For easy reference, we call this instantiation of MW-Net as \textbf{LogitNet}.
The structure of LogitNet is shown on the center of Figure \ref{fig:compare}.

%As shown in Figure 1, LogitNet would not produce polarized results, while it could still assign high sample weights to some noisydata. This is because LogitNet may contain too much redundant information with the unprocessed logits.

\subsection{MetaInfoNet: An Instatiation of MW-Net with Information Bottleneck}
%Information Bottleneck for Learning Representation}
Although LogitNet introduces more information to the weighting function, it also leads to a new problem: the unprocessed logits may contain too much redundant information for the weighting task. To ``squeeze out" the redundant information of the inputs for LogitNet, we propose a novel algorithm by information bottleneck (IB) principle \cite{alemi2016deep, shamir2010learning, tishby2000information}, called \textbf{MetaInfoNet}. Instead of directly using the output of classifier $\boldsymbol{z} = f_{\Theta}(\boldsymbol{x})$ in LogitNet, we aim to learn a representation $\boldsymbol{\psi}$ that captures the most relevant information from the logit $\boldsymbol{z}$ and removes redundant formation, with respect to the weighting task. In this way, the weight of sample $\boldsymbol{x}_i$ can be generated by $w_i=\text{MLP}(\boldsymbol{\psi}_i \odot \boldsymbol{e}(y_i))$.

\begin{figure}[!t]
\centering
\includegraphics[scale=0.3]{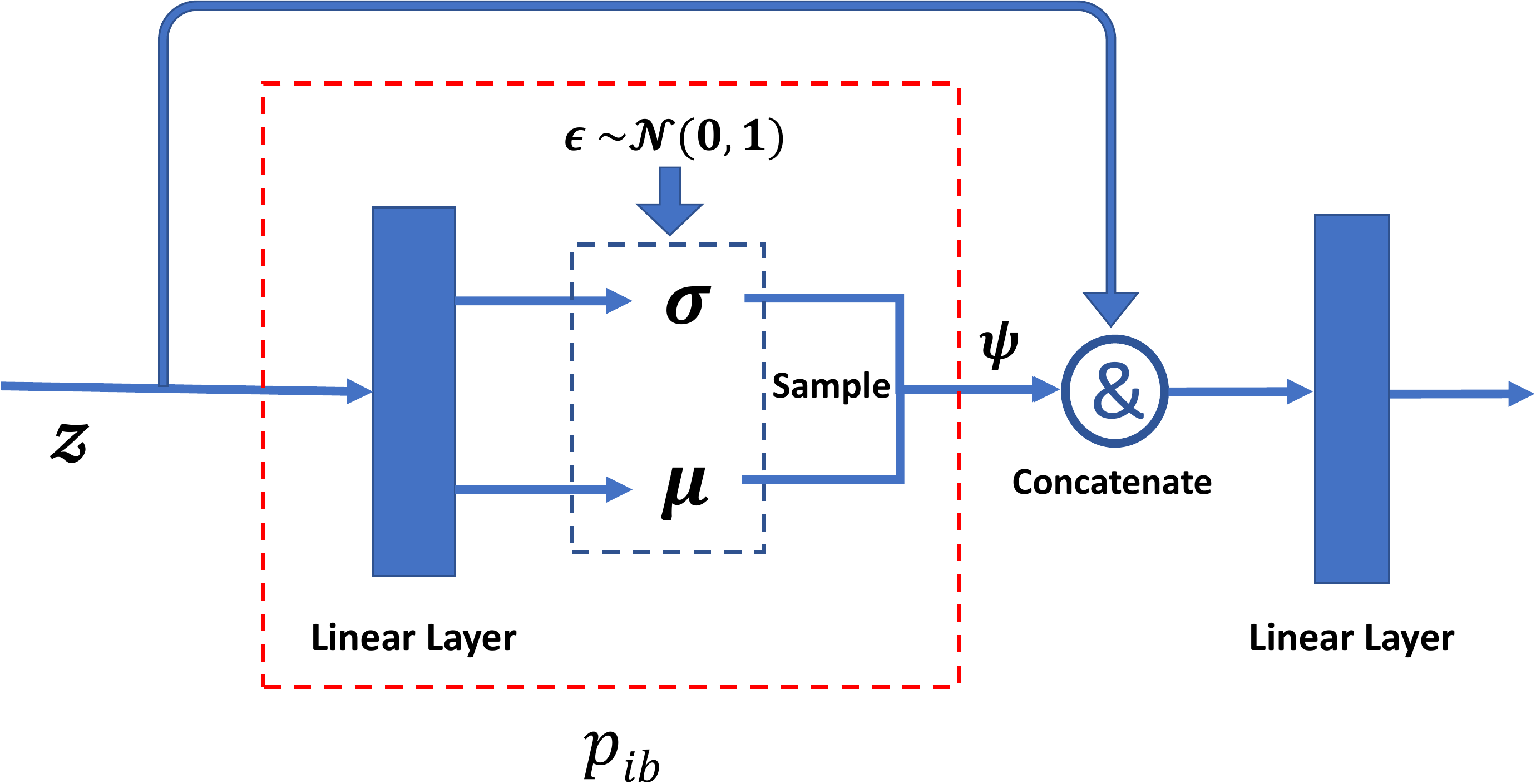}
\caption{The detailed structure of IB layer.}
\label{fig:MetaInfoNet}
\end{figure}

%To achieve this goal, we introduce the IB functional, which is built upon a tradeoff between the minimality of representation of input, and sufficiency of information on the weighting task.

Concretely, we propose an information-theoretic regularization by limiting the mutual information between the logit and the learned representation: $I(Z;\Psi) \leq I_c,$,
%todo: add introduction for I_c
where $Z$ is a random variable (for the logit vector) with a probability density function $p_{Z}(\boldsymbol{z})$, $\Psi$ is a random variable (for the learned representation vector) with a probability density function $p_{\Psi}(\boldsymbol{\psi})$ and $I_c$ is a constant that represents threshold of the mutual information. The constraint on mutual information can be viewed as a penalty term restricting the information from the logit vector to the learned representation vector.

Consequently, for learning an effective representation $\boldsymbol{\psi}$, we add the mutual information constraint for optimizing MW-Net:
\begin{gather}
\mathcal{L}^{\mathrm{meta}}(\Theta(\Phi)) = \sum\nolimits_{i=1}^M L^{\mathrm{meta}}_i(\Theta(\Phi)), \text{s.t.} \ I(Z;\Psi) \leq I_c,
\end{gather}

In practice, we propose to minimize the following objective using the information bottleneck Lagrangian:
\begin{align}
\sum\nolimits_{i=1}^M L^{\mathrm{meta}}_i(\Theta(\Phi)) + \lambda I(Z;\Psi),
\end{align}
where $\lambda$ is the Lagrange multiplier. As $\lambda$ increases, the information from $Z$ to $\Psi$ becomes denser while keeping relevant to the meta learning task. In the above equation, $I(Z;\Psi)$ is defined as:
\begin{align}
I(Z;\Psi) &=\iint p\left(\boldsymbol{z}, \boldsymbol{\psi}\right) \log \frac{p\left(\boldsymbol{z}, \boldsymbol{\psi}\right)}{p\left(\boldsymbol{z}\right) p\left(\boldsymbol{\psi}\right)} \mathrm{d} \boldsymbol{z} \mathrm{d} \boldsymbol{\psi} \\
&=\iint p\left(\boldsymbol{z}\right) p_{ib}\left(\boldsymbol{\psi} \mid \boldsymbol{z}\right) \log \frac{p_{ib}\left(\boldsymbol{\psi} \mid \boldsymbol{z}\right)}{p\left(\boldsymbol{\psi}\right)} \mathrm{d} \boldsymbol{z} \mathrm{d} \boldsymbol{\psi} \nonumber
\end{align}
where $p(\boldsymbol{z}, \boldsymbol{\psi})$ is the joint probability of $\boldsymbol{z}$ and $\boldsymbol{\psi}$, and $p_{ib}$ denotes the information bottleneck (IB) layer.

Unfortunately, computing the marginal distribution $p(\boldsymbol{\psi}) = \int p\left(\boldsymbol{\psi} \mid \boldsymbol{z}\right) p\left(\boldsymbol{z}\right) d \boldsymbol{z}$ is quite challenging as we do not know the prior distribution of $p(\boldsymbol{z})$. With the help of variational information bottleneck \cite{alemi2016deep}, we use a Gaussian approximation $h(\boldsymbol{\psi})$ of the marginal distribution $p(\boldsymbol{\psi})$ and view $p_{ib}$ as multivariate variational encoders (see Figure \ref{fig:MetaInfoNet}). Since $D_{\mathrm{KL}}\left[p\left(\boldsymbol{\psi}\right) \| h\left(\boldsymbol{\psi}\right)\right] \geq 0$, where the $D_{\mathrm{KL}}$ is the Kullback-Leibler divergence, we expand the KL term and get $\int p\left(\boldsymbol{\psi}\right) \log p\left(\boldsymbol{\psi}\right) d \boldsymbol{\psi} \geq \int p\left(\boldsymbol{\psi}\right) \log h\left(\boldsymbol{\psi}\right) d \boldsymbol{\psi}$, an upper bound on the mutual information $I(Z;\Psi)$ can be obtained via the KL divergence:
\begin{align}
\nonumber
I\left(Z ; \Psi\right) & \leq \int p\left(\boldsymbol{z}\right) p_{ib}\left(\boldsymbol{\psi} \mid \boldsymbol{z}\right) \log \frac{p_{ib}\left(\boldsymbol{\psi} \mid \boldsymbol{z}\right)}{h\left(\boldsymbol{\psi}\right)} d \boldsymbol{z} d \boldsymbol{\psi} \\
\nonumber
&=\mathbb{E}_{\boldsymbol{z} \sim p\left(\boldsymbol{z}\right)}\left[D_{\mathrm{KL}}\left[p_{ib}\left(\boldsymbol{\psi} \mid \boldsymbol{z}\right) \| h\left(\boldsymbol{\psi}\right)\right]\right],
\end{align}
which further provides an upper bound $\widetilde{\mathcal{L}}^{\mathrm{meta}}(\Theta(\Phi))$ on the regularized empirical objective that we minimize:
\begin{align}
\label{eq:kl_reg}
\widetilde{\mathcal{L}}^{\mathrm{meta}}(\Theta(\Phi)) & = \sum\nolimits_{i=1}^M L^{\mathrm{meta}}_i(\Theta(\Phi)) \\
\nonumber
& + \lambda \sum\nolimits_{j=1}^N\left[D_{\mathrm{KL}}\left[p_{ib}\left(\boldsymbol{\psi}_j \mid \boldsymbol{z}_j\right) \| h\left(\boldsymbol{\psi}_j\right)\right]\right].
\end{align}

Recall that the learned representation $\boldsymbol{\psi}$ via IB layer is used to generate the weight $w_i=\text{MLP}(\boldsymbol{\psi}_i \odot \boldsymbol{e}(y_i))$, then we optimize the MW-Net with the new meta loss in Eq. (\ref{eq:kl_reg}) to control the information capacity of the learned representation $\boldsymbol{\psi}$, while keeping the task-relevant information.

The network structure of the IB layer is shown in Figure \ref{fig:MetaInfoNet}. Specifically, a linear layer is applied to map logit $\boldsymbol{z}$ to a multivariate normal distribution. Then we use the ``re-parameterization" trick \cite{kingma2013auto} to sample a representation $\boldsymbol{\psi}$ from the learned distribution, parameterized by a mean vector $\mu$ and a variance vector $\sigma$. Another alternative trick is to implement skip connection by concatenating the representation $\boldsymbol{\phi}$ and the origin logit $\boldsymbol{z}$. In such a manner, we emphasize the most relevant information in $\boldsymbol{\psi}$, with respect to the learning task, while keeping intact information from the logit $\boldsymbol{z}$. To keep aligned with the dimension of label embedding for the element-wise product, we add a linear layer to scale the dimension of the learned representation $\boldsymbol{\psi}$. %The detailed structure of IB layer is presented in Figure \ref{fig:MetaInfoNet}.

%We encode the logits through an information bottleneck that serves to restrict information in the input to only the most relevant parts of the input, and allows the weighting network to focus on specific information in the input. 

\subsection{Convergence Analysis}
Here, we further provide the convergence analysis of MetaInfoNet. We show that MetaInfoNet converges to the critical points of both the meta and training loss function under some mild conditions in the following two theorems.
\begin{theo}
\label{theorem1}
Suppose the loss function $L$ is Lipshchitz smooth with constant $\rho$, and $\mathcal{V}(\cdot)$ is differential with a $\delta$-bounded gradients and twice differential with its Hessian bounded by $\mathfrak{B}$, and the loss function $L$ has $\gamma$-bounded gradients with respect to training/meta data. Let the learning rate $\alpha_t$ (for updating classifier) satisfies $\alpha_t=\min\{1,\frac{k}{T}\}$, for some constant $k>0$, such that $\frac{k}{T}<1$, and the learning rate $\beta_t$ (for updating meta weight net) satisfies $\beta_t=\min\{\frac{1}{L},\frac{c}{\sigma\sqrt{T}}\}$ for some constant $c>0$ where $\sigma$ is the variance of drawing uniformly mini-batch sample at random, such that $\frac{\sigma\sqrt{T}}{c}>\rho$ and $\sum_{t=1}^{\infty}\beta_t\leq\infty,\sum_{t=1}^{\infty}\beta_t^2\leq\infty$. Then we have
\begin{gather}
\nonumber
\lim\limits_{t\rightarrow\infty}\mathbb{E}\Big[\left\|\nabla_{\Theta}\mathcal{L}^{\mathrm{train}}(\Theta^{(t)};\Phi^{(t)})\right\|^2_2\Big] = 0.
\end{gather}
\end{theo}
\begin{theo}
\label{theorem2}
When the conditions in Theorem \ref{theorem1} holds, our proposed algorithm can achieve $\mathbb{E}[\left\|\nabla \mathcal{L}^{\mathrm{meta}}(\Theta^{(t)})\right\|_2^2]\leq\epsilon$ in $\mathcal{O}(1/\epsilon^2)$ steps, more specifically,
\begin{gather}
\nonumber
\min_{0\leq t\leq T}\mathbb{E}[\left\|\nabla_{\Theta} \mathcal{L}^{\mathrm{meta}}(\Theta^{(t)})\right\|_2^2]\leq\mathcal{O}(\frac{O}{\sqrt{T}}),
\end{gather}
\noindent where $C$ is some constant independent of the convergence process.
\end{theo}
It is worth noting that we have shown that our proposed method MetaInfoNet and LossNet \cite{shu2019meta} are different instantiations of MW-Net. Therefore, MetaInfoNet shares the same convergence properties as LossNet if we adopt the same assumptions for analyzing the two methods. We omit the proofs of the two theorems, as they are very similar to those proofs in \cite{shu2019meta}.
\section{Experiments}
In this section, we first implement experiments on simulated datasets with class imbalance and noisy labels. Then two real-world datasets with biased training data are also used to verify the effectiveness of our proposed algorithms.

\subsection{Experiment setup}
\noindent\textbf{Datasets.} For the class imbalance setting, we verify the effectiveness of our proposed algorithm on the Long-Tailed version of CIFAR10 and CIFAR100 datasets \cite{cui2019classbalancedloss}. Specifically, we reduce the number of training examples per class according to an exponential function $n = n_i\mu^i$, where $i$ is the class index, $n_i$ is the original number of training samples in the largest class divided by the smallest. Following LossNet \cite{shu2019meta}, we randomly select ten images per class in the validation set as the meta dataset.

For noisy label settings, we conduct experiments on simulated noisy datasets, including CIFAR10 and CIFAR100 \cite{krizhevsky2009learning}. We study two types of label noise on the training set: 1) \textbf{Flip noise.} Following LossNet \cite{shu2019meta}, we mainly consider the case that the label of each example is flipped to two similar classes with equal probability $\frac{p}{2}$ (Flip-2). 2) \textbf{Uniform noise.} The label of each example is changed to a random class with probability $p$ independently. To verify the effectiveness of our algorithms in various settings, we also compare our algorithms with existing meta reweighting algorithms in a harder case: One class is randomly selected as the similar class with probability $p$ (Flip-1). In these cases, 1000 images with clean labels in the validation set are randomly selected as the meta dataset.

To verify the effectiveness of the proposed method on real-world data, we conduct experiments on the Clothing1M dataset \cite{xiao2015learning}, containing 1 million images of clothing obtained from online shopping websites with 14 categories, e.g., T-shirt, Shirt, Knitwear. The labels are generated by using surrounding texts of the images provided by the sellers, and therefore contain many errors. We use the 7k clean data as the meta dataset. For preprocessing, we resize the image to $256 \times 256$, crop the middle $224 \times 224$ as input, and perform normalization. Moreover, we also compare our proposed algorithms with existing meta reweighting algorithms on ANIMAL10N dataset \cite{song2019selfie}, which contains 5 pairs of confusing animals with a total of 55,000 images. In our experiments, the training dataset contains 50,000 pictures with noisy labels, and 1000 pictures with clean labels are randomly selected as the meta dataset. The rest 4000 images are set as the test dataset.

\noindent\textbf{Network Structure and Optimizer.} For experiments on CIFAR10 and CIFAR100, we train ResNet-32 \cite{he2016deep} for the class imbalance settings. For the settings of label noise, we adopt WRN-28-10 \cite{zagoruyko2016wide} for uniform noise and ResNet-32 for flip noise. Specifically, SGD optimizer is applied with a momentum 0.9, a weight decay $5 \times 10^{-4}$, an initial learning rate 0.1, and the batch size is set to 100. The learning rate of the classifier network is divided by 10 after 80 epochs and 100 epochs (for a total of 120 epochs), and after 30 epochs and 40 epochs (for a total of 50 epochs). 

For experiments on Clothing1M, we use ResNet-50 pre-trained on ImageNet, following the previous works \cite{patrini2017making,shu2019meta,tanaka2018joint}. About the optimizer, we use SGD with a momentum 0.9, a weight decay $10^{-3}$, and an initial learning rate 0.01, and batch size 32. The learning rate of ResNet-50 is divided by 10 after 5 epochs (for a total 10 epochs). For experiments on ANIMAL10N, we also use ResNet-32. The setting of the SGD optimizer is the same as that on CIFAR10. We run 100 epochs in total, and the learning rate of the classifier network is divided by 10 after 50 epochs and 75 epochs. 

For the training of meta weighting network, we set the learning rate to $10^{-3}$ and set the weight decay to $5 \times 10^{-4}$. About the $\lambda$ in Eq. (\ref{eq:kl_reg}), we search it in $[0, 0.01, 0.03, 0.1, 0.3, 1]$. Additionally, we can set an interval parameter to control the updating of MW-Net, e.g., when the interval is 10, we update the MW-net every 10 iterations. We implement all methods with default parameters by PyTorch, and conduct all the experiments on NVIDIA Tesla V100 GPUs. We repeated the experiments 5 times with different random seeds for network initialization and label noise generation.

\noindent\textbf{Compared Methods.} For class imbalance setting, we compare our proposed algorithm with: 
1) Standard, which simply uses CE loss to train the DNNs with equal weights;
2) Focal loss \cite{lin2017focal}, which emphasizes harder examples by sample reweighting;
3) Class-balanced \cite{cui2019classbalancedloss}, which represents one of the state-of-the-arts of the predefined sample reweighting methods;
4) Fine-tuning, which finetunes the standard model on the meta dataset to further enhance its performance;
5) L2RW \cite{ren18l2rw}, which leverages an additional meta dataset to adaptively assign weights on training samples;
6) LossNet \cite{shu2019meta}, which automatically learns an explicit loss-weight function in a meta-learning manner.
For noisy label setting, the compared methods include: 
1) Standard;
2) Bootstrap \cite{reed2014training};
3) S-Model \cite{goldberger2016training};
4) SPL \cite{kumar2010self};
5) Focal Loss \cite{lin2017focal};
6) Co-teaching \cite{han2018co};
7) D2L \cite{ma2018dimensionality};
6) Fine-tuning;
7) MentorNet \cite{jiang2018mentornet};
8) L2RW \cite{ren18l2rw};
9) LossNet \cite{shu2019meta}.

\begin{table*}[!t]
\centering
\resizebox{1.0\textwidth}{!}{
\setlength{\tabcolsep}{1.5mm}{
\begin{tabular}{c|c|cccccc|cccccc}
\toprule
\multirow{2}{*}{--} &  \multirow{2}{*}{Algorithms} & \multicolumn{6}{c|}{Long-Tailed CIFAR-10}              & \multicolumn{6}{c}{Long-Tailed CIFAR-100}             \\ 
% \midrule
\cmidrule{3-14}&& 200        & 100        & 50        & 20        & 10        & 1     &  200        & 100        & 50        & 20        & 10        & 1       \\ 
\midrule
\multirow{6}{*}{Baselines} & Standard & 65.68 & 70.36 & 74.81 & 82.23 & 86.39 & 92.89 & 34.84 & 38.32 & 42.85 & 51.14 & 55.71 & 70.50             \\ 
&Focal loss & 65.29 & 70.38 & 76.71 & 82.76 & 86.66 & 93.03 & 35.62 & 38.41 & 44.32 & 51.95 & 55.78 & 70.52         \\ 
&Class-balanced          & 68.89 & 74.57 & 79.27 & 84.36 & 87.49 & 92.89 & 36.23 & 39.60 & 45.32 & 52.59 & 57.99 & 70.50          \\
\cmidrule{2-14}
& Fine-tuning          & 66.08 & 71.33 & 77.42 & 83.10 & 86.47 & \textbf{93.10} & \textbf{36.54} & 40.65 & 45.32 & 52.11 & 57.23 & \textbf{70.68}        \\ 
&L2RW & 66.51 & 74.16 & 78.93 & 82.12 & 85.19 & 89.25 & 33.38 & 40.23 & 44.44 & 51.64 & 53.73 & 64.11             \\ 
&LossNet & 66.37 & 72.92 & 77.26 & 83.36 & 86.98 & 92.83 & 33.15 & 40.24 & 44.67 & 52.73 & 56.63 & 68.91          \\ 
\midrule
\multirow{2}{*}{Ours} & LogitNet & 67.92 & 73.36 & 78.85 & 84.31 & 86.85 & 92.36 & 36.00 & 40.21 & 45.39 & \textbf{52.81} & 57.69 & 69.84          \\ 
&MetaInfoNet & \textbf{69.28} & \textbf{75.26} & \textbf{80.41} & \textbf{84.87} & \textbf{87.90} & 92.84 & 35.69 & \textbf{41.36} & \textbf{45.45} & 52.53 & \textbf{58.08} & 70.50          \\ 
\bottomrule                 
\end{tabular}
}
}
\caption{Test accuracy (\%) on long-tailed CIFAR-10 and CIFAR-100. The best results are highlighted in bold.}
\label{tab:imb_reults}
\end{table*}

\begin{table*}[!t]
    \centering
    \resizebox{1.0\textwidth}{!}{
    \setlength{\tabcolsep}{2mm}{
    \begin{tabular}{c|c|ccc|ccc}
\toprule
    \multicolumn{2}{c|}{Datasets}& \multicolumn{3}{c|}{CIFAR10}& \multicolumn{3}{c}{CIFAR100} \tabularnewline
    \midrule
    \multicolumn{2}{c|}{Noisy Rate} &\multicolumn{1}{c}{0\%} &\multicolumn{1}{c}{20\%} &\multicolumn{1}{c|}{40\%} &\multicolumn{1}{c}{0\%} &\multicolumn{1}{c}{20\%} &\multicolumn{1}{c}{40\%}
    \tabularnewline
    \midrule
    \multirow{11}{*}{Baselines} &Standard &92.89$\pm$0.32&76.83$\pm$2.30&70.77$\pm$2.31&70.50$\pm$0.12&50.86$\pm$0.27&43.01$\pm$1.16\tabularnewline
                            & Bootstrap  &92.31$\pm$0.25&88.28$\pm$0.36&81.06$\pm$0.76&69.02$\pm$0.32&60.27$\pm$0.71&50.40$\pm$1.01\tabularnewline    
                            &  S-Model &83.61$\pm$0.31&79.25$\pm$0.30&75.73$\pm$0.32&51.46$\pm$0.20&45.45$\pm$0.25&43.81$\pm$0.15\tabularnewline
                            & Self-paced &88.52$\pm$0.21&87.03$\pm$0.34&81.63$\pm$0.52&67.55$\pm$0.27&63.63$\pm$0.30&53.51$\pm$0.53\tabularnewline
                            & Focal Loss &93.03$\pm$0.61&86.45$\pm$0.19&80.45$\pm$0.97&70.02$\pm$0.53&61.87$\pm$0.30&54.13$\pm$0.40\tabularnewline      
                            & Coteaching &89.87$\pm$0.10&82.83$\pm$0.85&75.41$\pm$0.21&63.31$\pm$0.05&54.13$\pm$0.55&44.85$\pm$0.81\tabularnewline
                            & D2L   &92.02$\pm$0.14&87.66$\pm$0.40&83.89$\pm$0.46&68.11$\pm$0.26&63.48$\pm$0.53&51.87$\pm$0.33\tabularnewline
                            \cmidrule{2-8}
                            & Fine-tune &\textbf{93.23$\pm$0.23}&82.47$\pm$3.64&74.07$\pm$1.56&70.72$\pm$0.22&56.98$\pm$0.50&46.37$\pm$0.25\tabularnewline
                            &  MentorNet &92.13$\pm$0.30&86.36$\pm$0.31&81.76$\pm$0.28&70.24$\pm$0.21&61.97$\pm$0.47&52.66$\pm$0.56\tabularnewline
                            &  L2RW &89.25$\pm$0.37&87.86$\pm$0.36&85.66$\pm$0.51&64.11$\pm$1.09&57.47$\pm$1.16&50.98$\pm$1.55\tabularnewline
                            &  LossNet &91.99$\pm$0.54&89.25$\pm$0.38&85.31$\pm$0.19&68.46$\pm$0.17&64.53$\pm$0.41&57.89$\pm$0.79\tabularnewline
    \midrule
    \multirow{2}{*}{Ours} & LogitNet &91.75$\pm$0.23&88.52$\pm$0.46&85.23$\pm$0.38&69.11$\pm$0.31&64.10$\pm$0.56&56.79$\pm$0.85\tabularnewline
                            & MetaInfoNet &92.55$\pm$0.35&\textbf{90.07$\pm$0.43}&\textbf{86.63$\pm$0.47}&\textbf{70.81$\pm$0.33}&\textbf{65.70$\pm$0.50}&\textbf{58.09$\pm$0.37}\tabularnewline
    \bottomrule
    \end{tabular}}}
    \caption{Average test accuracy (\%) with standard deviation on synthesized datasets under Flip-2 noise (over 5 trials). The best results are highlighted in bold.}
    \label{tab:flip2}
\end{table*}

\subsection{Experiments with Class Imbalance}

Table \ref{tab:imb_reults} shows the test accuracy of different algorithms on long-tailed CIFAR10 and long-tailed CIFAR100. As we can see, our proposed algorithms work better than all the baselines in most of the long-tailed cases, which demonstrates their robustness against class imbalance. When imbalance factor is 1, e.g., there are the same number of samples in all classes, the fine-tuning method obtains the best performance on both CIFAR10 and CIFAR100, and our algorithms also get comparable performance. As the imbalance factor rises from 10 to 200, the datasets become more and more unbalanced and the accuracy of all algorithms decreases gradually. We observe that LogitNet achieves better or comparable results compared to LossNet in most cases, which indicates introducing more information can improve the effectiveness of MW-Net. Moreover, MetaInfoNet consistently performs better than LogitNet in most cases, verifying the effectiveness of the information bottleneck method while learning the representation.

\begin{table}[!t]
    \centering
    \resizebox{0.70\textwidth}{!}{
    \setlength{\tabcolsep}{2.5mm}{
   \begin{tabular}{c|c|c|c|c}
    \toprule
    % \multicolumn{2}{c|}{Datasets}& \multicolumn{3}{c}{CIFAR10} \tabularnewline
    % \midrule
    \multicolumn{2}{c|}{Noisy Rate} &\multicolumn{1}{c}{0\%} &\multicolumn{1}{c}{40\%} &\multicolumn{1}{c}{60\%} \tabularnewline
    \midrule
    \multirow{11}{*}{Baselines} &Standard &95.60$\pm$0.22&68.07$\pm$1.23&53.12$\pm$3.03\tabularnewline
                            & Bootstrap  &94.38$\pm$0.14&81.26$\pm$0.51&73.53$\pm$1.54\tabularnewline    
                            &  S-Model &83.79$\pm$0.11&79.58$\pm$0.33& 70.23$\pm$1.32\tabularnewline
                            & Self-paced & 90.81$\pm$0.34&86.41$\pm$0.29&53.10$\pm$1.78\tabularnewline
                            & Focal Loss &95.70$\pm$0.15&75.96$\pm$1.31&51.87$\pm$1.19\tabularnewline      
                            & Coteaching &88.67$\pm$0.25&74.81$\pm$0.34&73.06$\pm$0.25\tabularnewline
                            
                            & D2L   &94.64$\pm$0.33&85.60$\pm$0.13&68.02$\pm$0.41\tabularnewline
                            \cmidrule{2-5}
                            & Fine-tune &\textbf{95.65$\pm$0.15}&80.47$\pm$0.25&78.75$\pm$2.40\tabularnewline
                            &  MentorNet & 94.35$\pm$0.42&87.33$\pm$0.22&82.80$\pm$1.35\tabularnewline
                            &  L2RW &89.25$\pm$0.10&86.92$\pm$0.19&82.24$\pm$0.36\tabularnewline
                            &  LossNet &94.06$\pm$0.65&88.63$\pm$0.71&83.11$\pm$0.82\tabularnewline
    \midrule
    \multirow{2}{*}{Ours} & LogitNet &94.17$\pm$0.41&87.11$\pm$0.24&81.87$\pm$0.13\tabularnewline
                            & MetaInfoNet &94.73$\pm$0.17&\textbf{88.71$\pm$0.25}&\textbf{83.64$\pm$0.47}\tabularnewline
\bottomrule
\end{tabular}
}
}
\caption{Average test accuracy (\%) with standard deviation on CIFAR10 with WRN-28-10 under Unif noise (over 5 trials). The best results are highlighted in bold.}
\label{tab:unif}
\end{table}

\subsection{Experiments with Noisy Labels}

\begin{table}[!t]
\centering
\renewcommand\arraystretch{1.2}
\resizebox{0.60\textwidth}{!}{
\setlength{\tabcolsep}{0.45mm}{
\begin{tabular}{c|c|c|cc|cc}
\toprule
\multirow{2}{*}{Datasets}   & \multirow{2}{*}{Noisy Rate} &  -- & \multicolumn{2}{c|}{Baselines (Meta)}             & \multicolumn{2}{c}{Ours}             \\ 
\cmidrule{3-7} &    &-- &         L2RW &  LossNet   &  LogitNet  & MetaInfoNet        \\ 
\midrule
\multirow{4}{*}{CIFAR10} & \multirow{2}{*}{20\%} & best & 87.52 & 90.78 & 88.81  & \textbf{90.89}       \\ 
                                               & & last & 87.32 & 90.44 & 88.43  & \textbf{90.78}       \\
                                                \cmidrule{2-7}
                      & \multirow{2}{*}{40\%}& best&86.05 & 84.92 & 84.90 & \textbf{87.86}         \\ 
                                           &  & last&83.65 & 80.93 & 80.43 & \textbf{87.61}         \\ 
\midrule
\multirow{4}{*}{CIFAR100} & \multirow{2}{*}{20\%} & best & 57.18 & 64.13 & 64.2  & \textbf{65.09}        \\ 
                                               &  & last & 55.84 & 63.89 & 63.62  & \textbf{65.04}        \\
                                               \cmidrule{2-7}
                      & \multirow{2}{*}{40\%} & best& 46.73 & 50.48 & 51.81 & \textbf{52.92}         \\ 
                                           &  & last& 45.35 & 48.96 & 50.99 & \textbf{51.44}        \\ 
\bottomrule                         
\end{tabular}
}
}
\caption{Test accuracy (\%) on synthesized datasets under Flip-1 noise. The best results are highlighted in bold.}
\label{tab:flip1}
\end{table}

Table \ref{tab:flip2} shows the test accuracy of ResNet-32 averaged over 5 repetitions on both CIFAR10 and CIFAR100 under Flip-2 noise. We can observe that MetaInfoNet achieves the best performance across both datasets and all noise rates, indicating the effectiveness of our proposed architecture. When the noise rate is 0\%, MetaInfoNet also gets comparable results compared to the Fine-tune baseline while performs better than other meta-reweighting algorithms. Additionally, the results of LogitNet are similar to those of LossNet, which shows that simply introducing more information cannot improve the capacity of meta sample weighting in this case.

Table \ref{tab:unif} presents the test accuracy of WRN-28-10 averaged over 5 repetitions on CIFAR10 under Unif noise. It can be observed that all the meta weighting algorithms, including MetaInforNet, LogitNet, LossNet and L2RW, perform the best among all algorithms, which demonstrates their robustness. The results also indicate that uniform noise is a relatively simple biased setting for meta weighting algorithms. 

% \begin{figure}[!t]
% \centering
% \includegraphics[scale=0.22]{img/ablation.pdf}
% \caption{Comparison of Test Accuracy on CIFAR10 with ResNet32 between different architectures for MW-Net under various biased settings.}
% \label{fig:compare_cifar10}
% \end{figure}

Now we turn to a more difficult noisy setting, Flip-1 noise, that each example's label is flipped to one similar class with total probability $p$. As shown in Table \ref{tab:flip1}, \textit{Best} denotes the scores of the epoch where the validation accuracy is optimal, and \textit{Last} denotes the average accuracy over the last 10 epochs at the end of the training stage.
%In this setting, the sample number with noisy labels in one class are close to that with correct labels, which makes the loss no longer reliable. This phenomenon is explicitly shown in the Flip-1(40\%) label noise case.
As we can see, LossNet performs poorly even gets lower test accuracy than L2RW, while MetaInfoNet attains more than 8\% improvement over LossNet. Moreover, MetaInfoNet keeps the best performance in both CIFAR10 and CIFAR100 and all noise rates. Specifically, in the 20\% label noise cases, all four meta weighting algorithms could maintain almost the $best$ accuracy in the last 10 epochs. When it improves to the 40\% label noise cases, MetaInfoNet keeps the advantage while the other three algorithms' performance decrease dramatically after reaching the top. The result shows that MetaInfoNet is more robust to label noise compared to the other meta-weighting algorithms and has a better capacity to adapt to different biased settings in training data. 

% Figure \ref{fig:compare_cifar10} presents the performance comparison of different architectures for MW-Net under different label noise settings. From the figure, we can observe that MetaInfoNet performs the best among the three algorithms in all the label noise settings. Although LossNet achieves comparable result in the uniform case, it cannot keep up with our proposed algorithm in the two flip cases, especially Flip-1. The results show that MetaInfoNet has a better capacity to adapt to different biased settings in training data.

\subsection{Experiments on Real-world Datasets}

Table \ref{tab:clothing1m} shows the classification accuracy on the Clothing1M test set. Specifically, MetaInfoNet achieves the best accuracy, while LogitNet performs worse than LossNet. The results show that the IB layer is an efficient way to improving the meta-weighting framework with logits and labels as inputs.

\begin{table}[ht]
\renewcommand\arraystretch{1.2}
\centering
\resizebox{0.6\textwidth}{!}{
\setlength{\tabcolsep}{2.0mm}{
\begin{tabular}{c|c|c|c|c|c}
\toprule
\# & Method & Accuracy & \# & Method & Accuracy\\
\hline
\hline
1 & Standard &  68.74  & 5 & MLNT \cite{li2019learning} &  73.47\\
2 & Bootstrap &  69.12 & 6 & LossNet &  73.03\\
3 & S-Model &  69.84 & 7 & LogitNet &  72.10\\
4 & LCCN \cite{yao2019safeguarded} &  73.07 & 8 &MetaInfoNet &  \textbf{73.94}\\

\bottomrule
\end{tabular}
% \vspace*{pt}
}}
\caption{Classification accuracy (\%) on Clothing1M.}
\label{tab:clothing1m}
\end{table}

We also compare our proposed algorithms to the Finetune baseline and the other meta-weighting algorithms on the ANIMAL-10N dataset, as shown in Table \ref{tab:animal10n}. Note that the noise rate is about 8\% but it is difficult for even people to distinguish between categories, the performance gains of meta weighting baselines are not huge compared to the Finetune method. In this case, MetaInfoNet still obtains nearly 2\% improvement over the other methods on \textit{best}. After all epochs, MetaInfoNet maintains this advantage over the best baseline method.

\begin{table}[ht]
\renewcommand\arraystretch{1.0}
\centering
\resizebox{0.60\textwidth}{!}{
\setlength{\tabcolsep}{9.5mm}{
\begin{tabular}{c|c|c}
\toprule
Method & Best & Last \\
\midrule
 Finetune &  83.3 & 82.21 \\
 L2RW & 82.17 & 81.32\\
 LossNet &  83.73 & 82.07\\
 LogitNet &  83.58 & 82.05\\
 MetaInfoNet &  \textbf{85.13} & \textbf{83.95}\\
\bottomrule
\end{tabular}
% \vspace*{pt}
}}
\caption{Classification accuracy (\%) on ANIMAL-10N.}
\label{tab:animal10n}
\end{table}

\section{Conclusion}

In this paper, we propose an effective approach called MetaInfoNet to improve the robustness of deep neural networks under various biased settings in training data. Compared with current meta reweighting algorithms that directly map loss to weight for each sample, MetaInfoNet could automatically learn effective representations as inputs for the meta weighting network by emphasizing task-related information with an information bottleneck strategy.
The Empirical results on simulated and real-world datasets demonstrate that the robustness of deep models trained by our proposed approach is superior to many state-of-the-art approaches in general biased settings, such as class imbalance, label noise, and more complicated real cases.
\newpage

\bibliographystyle{ieeetr}
\bibliography{arxiv-template}

\newpage
\end{document}